# Revealing the CO$_2$ emission reduction of ridesplitting and its determinants based on real-world data

**Wenxiang Li [a,b], Yuanyuan Li [a], Ziyuan Pu [c], Long Cheng [d], Lei Wang [e,*], Linchuan Yang [f,*]**

[a] Business School, University of Shanghai for Science and Technology, 516 Jungong Road, Shanghai 200093, P.R. China

[b] Center for Supernetworks Research, University of Shanghai for Science and Technology, 334 Jungong Road, Shanghai 200093, P.R. China

[c] School of Engineering, Monash University, Jalan Lagoon Selatan, 47500 Bandar Sunway, Malaysia

[d] School of Transportation, Southeast University, 2 Southeast University Road, Nanjing 211189 P.R. China

[e] College of Transport and Communications, Shanghai Maritime University, Shanghai 201306 China.

[f] Department of Urban and Rural Planning, Southwest Jiaotong University, Chengdu 611756, P.R. China

[*] = Corresponding author
E-mail addresses: wangleicuail@gmail.com (L. Wang); yanglc0125@swjtu.edu.cn (L. Yang)
Declarations of interest: none

# Abstract

Ridesplitting, which is a form of pooled ridesourcing service, has great potential to alleviate the negative impacts of ridesourcing on the environment. However, most existing studies only explored its theoretical environmental benefits based on optimization models and simulations. By contrast, this study aims to reveal the real-world emission reduction of ridesplitting and its determinants based on the observed data of ridesourcing in Chengdu, China. Integrating the trip data with the COPERT model, this study calculates the $CO_2$ emissions of shared rides (ridesplitting) and their substituted single rides (regular ridesourcing) to estimate the $CO_2$ emission reduction of each ridesplitting trip. The results show that not all ridesplitting trips reduce emissions from ridesourcing in the real world. The $CO_2$ emission reduction rate of ridesplitting varies from trip to trip, averaging at 43.15g/km. Then, interpretable machine learning models, gradient boosting machines, are applied to explore the relationship between the $CO_2$ emission reduction rate of ridesplitting and its determinants. Based on the SHapley Additive exPlanations (SHAP) method, the overlap rate and detour rate of shared rides are identified to be the most important factors that determine the $CO_2$ emission reduction rate of ridesplitting. Increasing the overlap rate, the number of shared rides, average speed, and ride distance ratio while decreasing the detour rate, actual trip distance, and ride distance gap can increase the $CO_2$ emission reduction rate of ridesplitting. In addition, nonlinear effects and interactions of the determinants are examined through the partial dependence plots. To sum up, this study provides a scientific method for the government and ridesourcing companies to better assess and optimize the environmental benefits of ridesplitting.

**Word count:** 7900 (excluding title, author names and affiliations, keywords, abbreviations, table/figure captions, acknowledgements and references)

**Highlights**

- The real-world $CO_2$ emission reductions of ridesplitting are quantified based on ridesourcing trip data.
- Gradient boosting machines are used to explore the determinants of $CO_2$ emission reductions of ridesplitting.
- The average $CO_2$ emission reduction rate of ridesplitting is 43.15g/km.
- 15% of ridesplitting trips even increase $CO_2$ emissions compared with regular ridesourcing.
- The overlap rate and detour rate of shared rides are the most important determinants.

# Nomenclature

| | |
|---|---|
| *ts* | Timestamp of the trajectory point |
| *t* | Travel time of the trajectory segment |
| *d* | Travel distance of the trajectory segment |
| *v* | Average travel speed of the trajectory segment |
| *T* | Total trajectory duration of the ridesourcing trip |
| *D* | Total trajectory distance of the ridesourcing trip |
| *NSR* | Number of shared rides in the ridesplitting trip |
| *TT* | Trip time |
| *TD* | Trip distance |
| *C* | Calibration coefficient |
| *EF* | $CO_2$ emission factor |
| *E* | $CO_2$ emission |
| *OG* | Origin grid |
| *DG* | Destination grid |
| $\widehat{TD}$ | Median (baseline) trip distance of substituted single rides |
| $\widehat{E}$ | Median (baseline) $CO_2$ emission of substituted single rides |
| *SD* | Travel distance saved by the ridesplitting trip |
| *ER* | $CO_2$ emission reduction of the ridesplitting trip |
| *ERP* | $CO_2$ emission reduction percentage of the ridesplitting trip |
| *ERR* | $CO_2$ emission reduction rate of the ridesplitting trip |
| *OD* | Overlap distance of the ridesplitting trip |
| *OT* | Overlap time of the ridesplitting trip |
| *OR* | Overlap rate of the ridesplitting trip |
| *DD* | Detour distance of the ridesplitting trip |
| *DR* | Detour rate of the ridesplitting trip |
| *Φ* | A finite set of overlapping trajectory segments of the ridesplitting trip |
| *Ω* | A finite set of all trajectory segments of the ridesplitting trip |
| *x* | Input of machine learning models |
| *∅* | SHAP value |
| *I* | Global importance value of the feature |
| *M* | Number of samples |

## Subscripts

| | |
|---|---|
| *i* | Trajectory segment |
| *k* | Single ride (regular ridesourcing trip) |
| *n* | Shared ride |
| *s* | Ridesplitting trip |
| *start* | Start timestamp |
| *end* | End timestamp |
| *j* | Feature for machine learning models |
| *m* | Sample for machine learning models |
| *S, C* | Subset of features |

# 1 Introduction

Under the global goal for climate change set in the Paris Agreement, many countries have announced to achieve carbon neutrality by the mid-21st century. As the main source of greenhouse gases, the transport sector emitted 24.6% of global $CO_2$ emissions from fuel combustion in 2020 [1]. It is also the most challenging sector to achieve carbon peak and neutrality, given its rapidly growing trend [2,3]. Therefore, it is imperative to explore pathways toward low-carbon transport.

The development of sharing economy has given rise to the widespread adoption of ridesourcing, which is an on-demand transportation service connecting drivers and passengers via smartphone applications [4]. It is also known as ride-hailing, e-hailing, and Transportation Network Company (TNC) services, such as Uber and Lyft in the U.S., Didi Express in China, and Ola in India [5]. On the one hand, ridesourcing can increase vehicle occupancy and reduce the cruising time and waiting time compared with traditional taxis [6–8]. On the other hand, ample evidence shows that it also increases vehicle kilometers traveled (VKT), traffic congestion, and emissions due to the induced travel demand and mode shifts from public transit, walking, and biking [9–11].

To alleviate the negative impact of ridesourcing, a high occupancy option called ridesplitting (or ride-pooling in some contexts) is provided on the ridesourcing platform. It enables ridesourcing passengers on similar routes to be matched in the same vehicle to share the trip and split the cost. Some studies have theoretically proved that ridesplitting has great potential to improve transport efficiency [12–14] and reduce traffic emissions [15–19] compared with regular ridesourcing. However, most existing studies are based on optimization models, ideal scenarios, or simulations [19]. In these studies, the environmental benefits of ridesplitting may be overestimated since the heterogeneity and uncertainty in the real-world situation are neglected. In reality, not all ridesplitting trips reduce emissions from ridesourcing due to the possible detours of shared rides. Therefore, the real-world emission reduction of ridesplitting and its determinants remain unrevealed.

To gain insight into the environmental impact of ridesplitting, this study aims to answer the following questions: 1) To what extent can ridesplitting reduce $CO_2$ emissions from ridesourcing trips in the real world? 2) What are the factors that affect the $CO_2$ emission reduction of ridesplitting? To this end, an empirical study is conducted based on observed data of ridesplitting trips in Chengdu, China. Integrating the trip data with the COPERT model, the $CO_2$ emissions of both single rides (regular ridesourcing) and shared rides (ridesplitting) are calculated. Then, the $CO_2$ emission reduction of each ridesplitting is quantified by subtracting the emissions of shared rides from the total emission of their substituted single rides. Furthermore, regression analysis of gradient boosting machines is used to explore the determinants of $CO_2$ emission reduction of ridesplitting. The contributions of this study can be summarized as follows:

- Instead of evaluating the potential (theoretical) environmental benefits of ridesplitting based on optimization models or simulations (as most previous studies do), this study assesses the actual (practical) environmental benefits based on the observed data of ridesplitting services in the real world.
- Instead of analyzing the overall (system) impact of ridesplitting on transportation and the environment (as most previous studies do), this study quantifies the individual effects of each ridesplitting trip on the VKT and emissions compared with its substitute.
- Instead of using traditional linear regression for modeling the relationship between independent and dependent variables, this study employs interpretable machine learning, namely gradient boosting machines, to identify the most important factors that influence emission reduction of ridesplitting. In addition, the nonlinear effects and interaction of these determinants can also be inferred.

The remainder of this study is organized as follows. Section 2 presents a review of relevant studies on the impact of ridesplitting. The data and preliminary analysis for this study are introduced in Section 3. Section 4 elaborates the methods for analyzing the emission reduction of ridesplitting. The results are analyzed and discussed in Section 5. The main findings and policy implications are summarized in the last section.

## 2 Literature review

This section presents the summary of existing studies related to the impacts of ridesplitting (ride-pooling) as well as on-demand (dynamic) ridesharing, which has a similar scope to ridesplitting [20,21]. Most of these studies focus on the improvement in traffic efficiency and the associated environmental benefits of ridesplitting compared with its substituted modes.

The impacts of ridesplitting on traffic efficiency mainly include the reductions in VKT, traffic congestion, fleet size, and travel time [22]. Zhang et al. [23] designed a ridesplitting recommendation system for taxicab services and found that it can reduce 60% of the total mileage, 41% of the passenger's waiting time, and 28% of the total travel time compared with its baseline. Santi et al. [12] proposed a novel method of shareability networks to quantify the benefits of vehicle pooling in taxi services. Based on the simulation of millions of taxi trips in New York City, they found that shared taxis can reduce vehicle fleets and accumulative trip distances by over 40%. Similarly, Alonso-Mora et al. [24] also showed that on-demand high-capacity ridesharing can significantly decrease the fleet size of taxis in New York City. Chen et al. [25] proved that dynamic ride pooling can averagely reduce the total travel distance by 18% and reduce the total number of vehicles by 32% based on real-world ride request data from three US cities. Chen et al. [26] analyzed the impact of ridesplitting on urban mobility based on real-world ridesourcing data and questionnaires in Hangzhou, China. In their study, the total VKT reduction of ridesplitting is estimated to be 58,124 km per day in Hangzhou, considering the modal shift from passenger/private vehicles to Hitch or DiDi Express ridesplitting. There is also evidence showing that ridesplitting can only reduce aggregate VKT by 8.21% in a mid-sized city, Haikou, which is much lower than that in megacities [19]. To sum up, all the above studies proved that ridesplitting has great potential to improve urban transportation efficiency. However, most of them focus on the overall effects of ridesplitting on system-wide traffic parameters [27], such as total VKT, total number of vehicles, and total travel time. In the real world, not all ridesplitting trips can reduce travel distance since ridesplitting may inevitably cause detours for picking up and dropping off other shared riders. Therefore, the individual effect of each ridesplitting trip and its determinants should be further studied.

The impacts of ridesplitting on the environment mainly include reductions in energy consumption, $CO_2$ emissions, and air pollutants. Yu et al. [15] evaluated both the direct and indirect environmental benefits of ridesharing through life cycle analysis of vehicle fuels and environmental input-output analysis in Beijing, including the energy savings and emission reduction caused by travel mode shift in the short term and attitude change towards car ownership in the mid and long term. Yin et al. [16] developed a comprehensive land-use transport model to analyze the impacts of several ride-sharing scenarios on $CO_2$ emissions in the Paris region, considering the rebound effects on travel behaviors. Cai et al. [17] found that ridesharing of 12083 taxis at a tolerance level of 10 min can reduce 28.3 million gallons of gasoline, 186 tons VOC, 199 tons $NO_x$, 53 tons $PM_{10}$, 25 tons $PM_{2.5}$, and 2392 tons CO emissions annually. Using taxi data in Shanghai as a proxy of the ridesharing demand, Yan et al. [18] inferred that ridesharing can provide 15-23% energy-saving and emission reductions in different scenarios based on the shareability network and COPERT model. Zhang et al. [28] assessed the emission reduction potential of ridesharing in Tokyo by assuming that travelers who originally used private cars and public transit will turn to ridesharing. Akimoto et al. [29] predicted the global energy consumption and emission reduction of the ride and car-sharing associated with fully autonomous cars in the future under the 2 °C or 1.5 °C targets. To sum up, all the above studies proved that ridesplitting can provide a substantial improvement in the environment. However, most of these studies quantified the potential (theoretical) environmental benefits of ridesplitting in hypothetical or optimal scenarios based on simulation or optimization models. However, the actual (practical) environmental benefits of ridesplitting in the real world have been rarely studied.

To the best of our knowledge, the only exception is the work of Li et al. [20]. They analyzed the spatially fine-grained emission reduction of ridesplitting services based on observed ridesplitting data in the real world. However, they only estimated the aggregate emission reductions per ride-km by comparing the emission factors of both regular ridesourcing and ridesplitting in the spatial grids. Their results neglect the heterogeneity of ridesplitting trips. In fact, whether a specific ridesplitting trip can reduce emissions depends on the travel-related attributes of shared rides and their substitutes. Therefore, the factors that influence the emission reduction of each ridesplitting trip are still unveiled. To fill these research gaps, this study aims to reveal the actual quantity of emission reduction of ridesplitting trips and its determinants based on the observed ridesplitting data in the real world.

## 3 Data preparation

### 3.1 Data and study area

The data used in this study include the ride order and GPS trajectory datasets of ridesourcing services of Didi Chuxing in Chengdu, China, and it is provided by the GAIA program [4]. The structure of the ride order dataset is shown in Table 1. It contains fields such as identity (ID), start and end times, pick-up and drop-off locations of each ride order. The structure of the GPS trajectory dataset is shown in Table 2. It contains fields such as driver ID, order ID, timestamp, longitude, and latitude of each GPS point, with an average sampling interval of 3 seconds.

**Table 1 Ride order data structure.**

| Field | Type | Sample | Comment |
|---|---|---|---|
| OrderID | String | mjiwdgkqmonDFvCk3ntBpron5mwfrqvI | Anonymized |
| Ride Start Time | String | 1501581031 | Unix Timestamp, in seconds |
| Ride End Time | String | 1501582195 | Unix Timestamp, in seconds |
| Pick-up Longitude | String | 104.11225 | GCJ-02 Coordinate System |
| Pick-up Latitude | String | 30.66703 | GCJ-02 Coordinate System |
| Drop-off Longitude | String | 104.07403 | GCJ-02 Coordinate System |
| Drop-off Latitude | String | 30.6863 | GCJ-02 Coordinate System |

**Table 2 GPS trajectory data structure.**

| Field | Type | Sample | Comment |
|---|---|---|---|
| DriverID | String | glox.jrrlltBMvCh8nxqktdr2dtopmlH | Anonymized |
| OrderID | String | jkkt8kxniovIFuns9qrrlvst@iqnpkwz | Anonymized |
| Timestamp | String | 1501584540 | Unix Timestamp, in seconds |
| Longitude | String | 104.04392 | GCJ-02 Coordinate System |
| Latitude | String | 30.6863 | GCJ-02 Coordinate System |

The ride order dataset covers over 6 million ridesourcing trips (including both single rides and shared rides) in Chengdu from November 1st to November 30th, 2016. However, the GPS trajectory dataset only covers a rectangular region with longitudes from 104.04 to 104.12E and latitudes from 30.65 to 30.72N in the GCJ-02 Coordinate System, which is also the central city area of Chengdu [30]. Therefore, the study area is this region, as shown in Fig.1. It is further divided into 289 grids of 500 m × 500 m for spatial aggregation of ridesourcing trips. Refer to previous studies for more details [4,20].

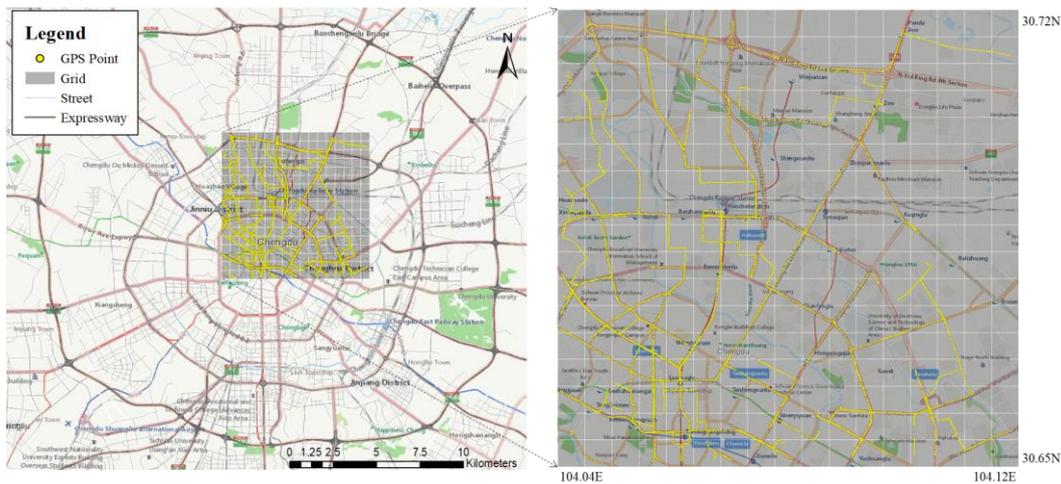

**Fig. 1. Study area and trajectory data samples**

## 3.2 Data processing

Although the ride order dataset contains whole trips across the city, the trajectory data of those trips were only provided in the study area. To focus on trips with complete trajectories, we first select the rides with both origins and destinations in the study area. Meanwhile, the trajectory data of selected rides in the study area are also selected. The selected ride order and trajectory data are then processed to extract the trip characteristics for further analysis, as shown in Fig.2. Some key features are highlighted in red in the figure.

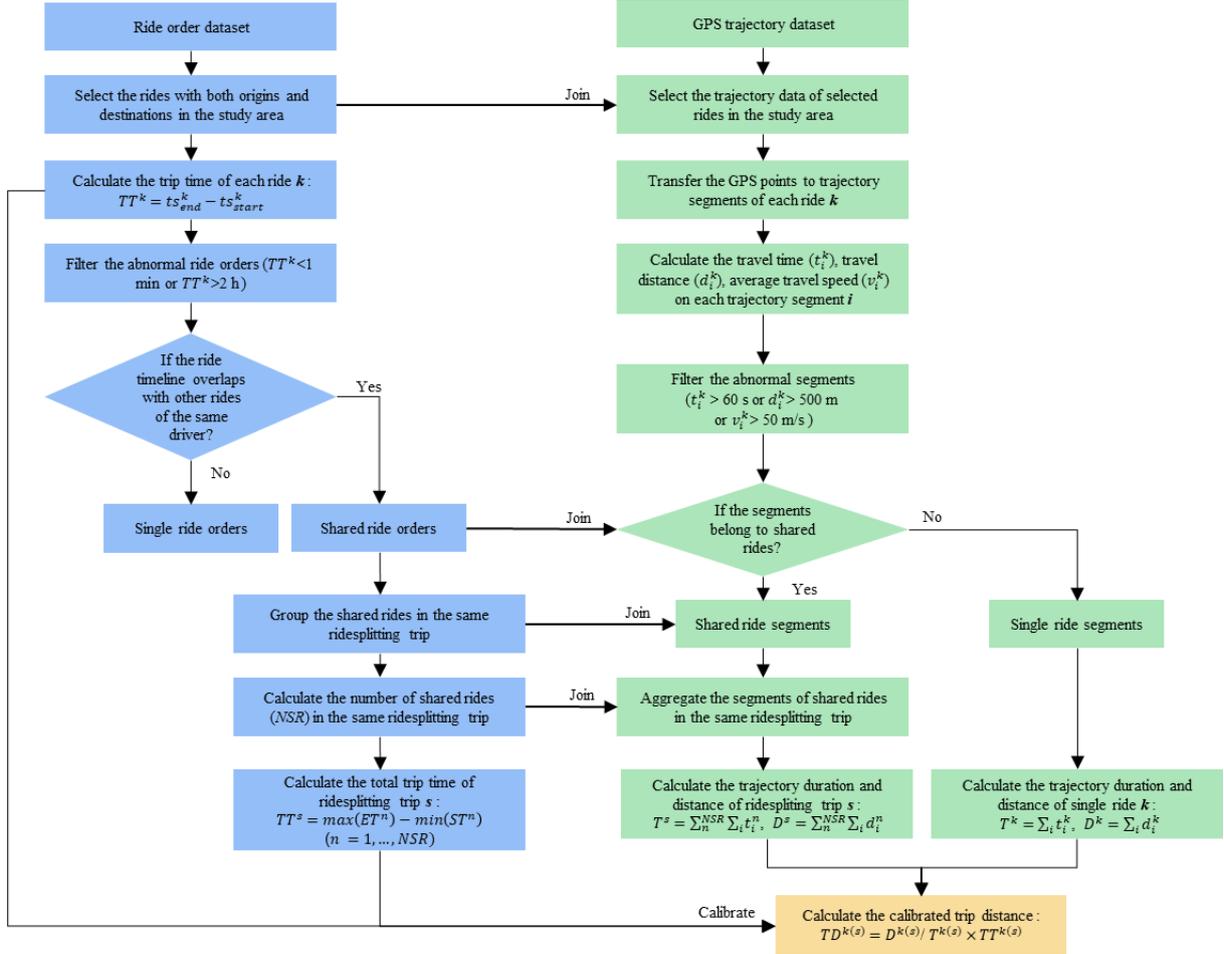

**Fig. 2. Flow chart of data processing.**

Based on the ride order data, the trip time ($TT^k$) of each ride $k$ is calculated by Eq (1) and used to filter the abnormal ride orders with $TT^k$<1 min or $TT^k$>2 h. Since the information about ridesplitting is not provided directly in the original order dataset, we adopt the ridesplitting trip identification algorithm proposed in our previous studies [4,20] to separate the shared rides from single rides. Then the shared rides in the same ridesplitting trip can be grouped and counted. For each ridesplitting trip $s$, it starts when the first passenger is picked up and ends when the last passenger is dropped off. Based on this, the total trip time of each ridesplitting trip $s$ can be calculated as Eq (2).

Meanwhile, the GPS points of each driver are transferred to trajectory segments and sorted by the timestamp. Then the travel time ($t_i^k$), travel distance ($d_i^k$), average travel speed ($v_i^k$) of each trajectory segment $i$ can be calculated. Considering the problems of signal drift and GPS data missing, the abnormal segments with $t_i^k > 60$ s or $d_i^k > 500$ m or $v_i^k > 50$ m/s are filtered to avoid inaccuracy of distance calculation. If the segments belong to shared rides in ridesplitting trip $s$, they will be aggregated to construct the complete trajectory of ridesplitting trip $s$. For each ridesplitting trip $s$ or single ride $k$, the total trajectory duration ($T$) and distance ($D$) can be calculated by summing the travel time and distance of all the segments of this trip, as shown in Eq (3)-(4). However, it should be noted that the trajectory distance is not necessarily equal to the trip distance because of the filtration of abnormal segments previously. Thus, the actual trip distance is calibrated using the trip time ($TT^k$) which is derived from ride order data and considered to be accurate, as shown in Eq (5)-(6).

$$TT^k = ts_{end}^k - ts_{start}^k \tag{1}$$

$$TT^s = max(ts_{end}^n) - min(ts_{start}^n), \quad n = 1, 2, ..., NSR \tag{2}$$

$$T^s = \sum_n^{NSR} \sum_i t_i^n, \quad T^k = \sum_k t_i^k \tag{3}$$

$$D^s = \sum_n^{NSR} \sum_i d_i^n, \quad D^k = \sum_k d_i^k \tag{4}$$

$$C^{k(s)} = TT^{k(s)} / T^{k(s)} \tag{5}$$

$$TD^{k(s)} = D^{k(s)} \times C^{k(s)} \tag{6}$$

where superscript $k$ denotes each ride order (both single ride and shared ride), superscript $s$ denotes each ridesplitting trip consisting of multiple shared rides, superscript $n$ denotes each shared ride in ridesplitting trip $s$, $NSR$ is the number of shared rides in ridesplitting trip $s$, subscript $i$ denotes each trajectory segment of ride $k$, subscript $start$ and $end$ denote departure and arrival time of each trip, $ts$ is the timestampe of the trajectory point, $TT$ is the actual trip time, $T$ is the total trajectory duration, $t$ is the travel time on the segment, $TD$ is the actual trip distance, $D$ is the total trajectory distance, $d$ is the travel time on the segment, $C$ is the calibration coefficient for trip distance and emissions. The above data and parameters are prepared for calculating the $CO_2$ emission reduction of ridesplitting and the possible explanatory variables.

### 3.3 Preliminary analysis

After the filtering of outside and abnormal data, 1.6 million ride orders are selected in the study area, including 1.5 million single rides and 0.1 million shared rides. Given the start time of each ride order in the study area, we can analyze and compare the temporal distributions of single rides and shared rides, as shown in Fig. 3. The blue and orange dots in the figure represent the percentage of hourly trips on weekdays and weekends, respectively. The error bars in the figure indicate the variation on different days. It can be concluded that the temporal patterns of single rides and shared rides are obviously different. For the single rides, there are three peaks at 8:00-10:00, 13:00-15:00, and 17:00-18:00 on weekdays but only two peaks (noon peak and evening peak) on weekends. For the shared rides, there is a salient peak at 14:00-15:00 on both weekdays and weekends. In addition, the ridership of shared rides is still high at 22:00-23:00 on weekends. These differences between single rides and shared rides indicate that ridesplitting mainly serves non-commuting trips during off-peak hours of traffic. This is consistent with previous studies [4].

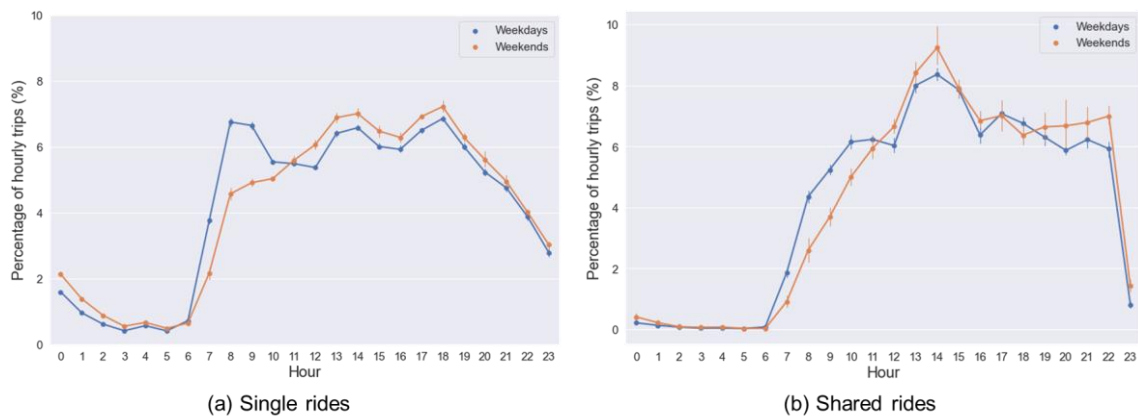

**Fig. 3 Temporal distributions of single rides and shared rides.**

By aggregating the origin and destination of each ride order into the 500m×500m grid, the spatial distribution of single rides and shared rides are presented, as shown in Fig.4. It shows that the spatial patterns of single rides and shared rides are very similar. The hot spots of trips are located near Tianfu Square, Chunxi Road, South Renmin Road (metro station), and East areas between 1st Ring Road and 2nd Ring Road. The shared rides are even denser in places with lots of shopping malls. It also indicates that people use ridesplitting more likely for entertainment-related activities, such as shopping or catering.

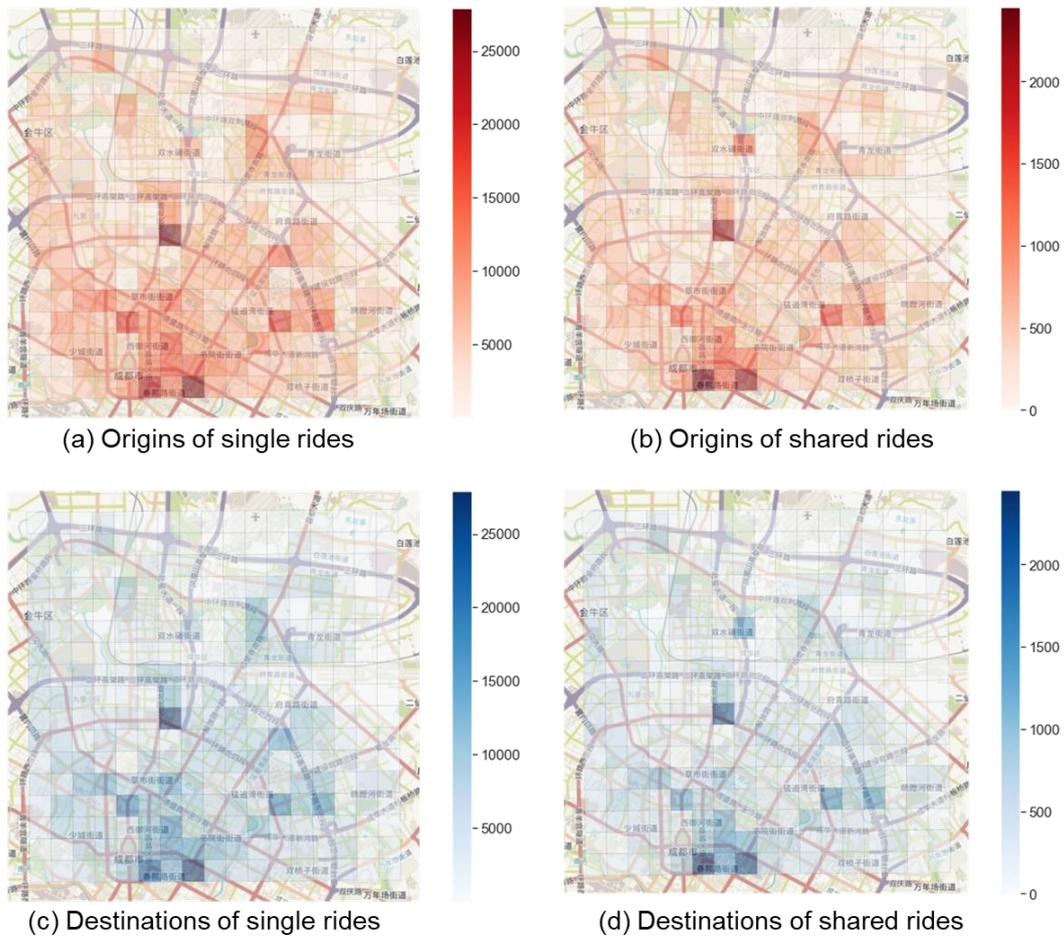

(a) Origins of single rides
(b) Origins of shared rides
(c) Destinations of single rides
(d) Destinations of shared rides

**Fig. 4 Spatial distributions of single rides and shared rides.**

## 4 Methodology

### 4.1 Calculation of the $CO_2$ emission of ridesourcing trips

Referring to many previous studies on vehicle emissions in China [31–33], this study adopts the COPERT model to calculate the emissions of ridesourcing trips in Chengdu. This model is designed for motor vehicles complying with the European emission standard. Since the emission standard implemented in China is similar to the European emission standard, the COPERT model is also applicable for this study. According to references [20,32], most ridesourcing vehicles in Chengdu in 2016 must meet China stage IV emission standard, which is equivalent to Euro 4 emission standard. In addition, the market share of electric cars in 2016 is still very low for Chengdu. Without loss of generality, this study assumes all ridesourcing vehicles to be small passenger cars fueled with petrol, complying with Euro 4 in the COPERT model [32].

The COPERT model calculates vehicle emissions based on experimentally obtained emission factors. These emission factors are speed-dependent and differ by fuel, vehicle class, and engine technology. The $CO_2$ emission factor of each trajectory segment $i$ can be calculated as Eq (7), and the $CO_2$ emission of the segment is calculated as Eq (8). Based on the data processing above, the $CO_2$ emission of a regular ridesourcing trip $k$ is equal to the sum of all the emissions of segments composing this trip, as shown in Eq (9). For ridesplitting trip $s$, the $CO_2$ emission of the whole trip is the sum of emissions from all shared rides in this trip, as shown in Eq (10).

$$EF_i = \left(\alpha \times v_i + \beta \times v_i + \gamma + \delta / v_i \right) / \left(\varepsilon \times v_i^2 + \zeta \times v_i + \eta \right) \qquad (7)$$

$$E_i = EF_i \times d_i \qquad (8)$$

$$E^k = C^k \times \sum_i E_i^k \qquad (9)$$

$$E^s = C^s \times \sum_n^{NSR} \sum_i E_i^n \qquad (10)$$

where $EF_i$ is the $CO_2$ emission factor on trajectory segment $i$ (unit: g/km), $v_i$ is the average travel speed on segment $i$ (unit: km/h), $\alpha$, $\beta$, $\gamma$, $\delta$, $\varepsilon$, $\zeta$, and $\eta$ are calibrated parameters for small petrol passenger cars complying with the Euro 4 emission standard, their values can be found in references [20,34], $E_i$ is the $CO_2$ emission on trajectory segment $i$, $d_i$ is the on trajectory segment $i$, $E^k$ is the $CO_2$ emission of regular ridesourcing trip $k$, $E^s$ is the $CO_2$ emission of ridesplitting trip $s$.

## 4.2 Estimation of the $CO_2$ emission reduction of ridesplitting trips

To analyze whether a ridesplitting trip reduces $CO_2$ emission compared with regular ridesourcing trips, the substituted single rides for shared rides should be found. If both the origin and destination of a single ride are matched with those of a shared ride, then the single ride is regarded as the substitute for the shared ride when ridesplitting is not available. However, matching the single rides and shared rides with exactly the same locations is difficult. To simplify this problem, the origins and destinations of both the single rides and shared rides are aggregated in the 500m×500m grids for approximatively matching, as shown in Fig. 5.

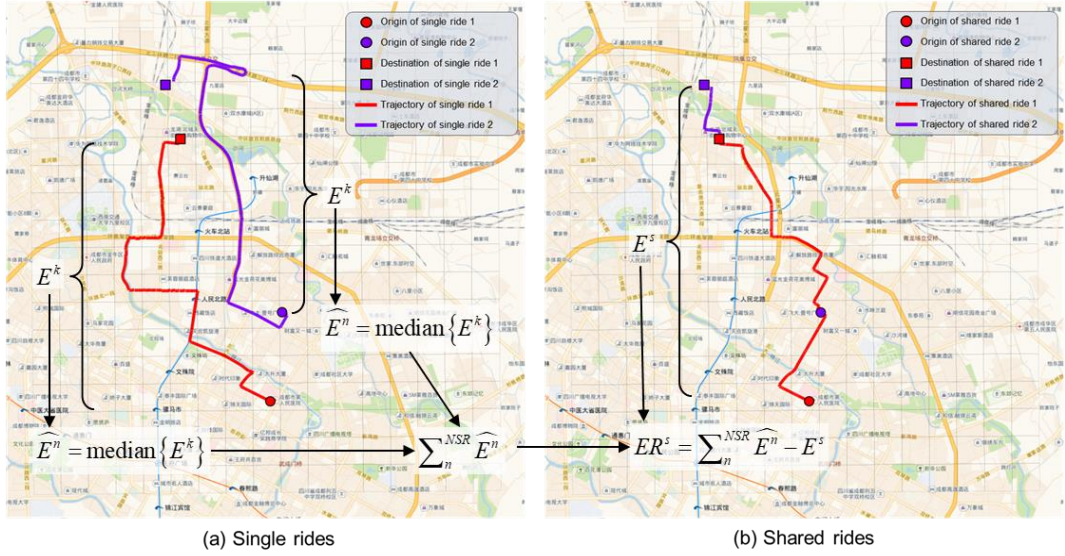

**Fig. 5 Example of substituted single rides for shared rides**

For example, there is a ridesplitting trip with two shared rides in Fig.5(b). We can find that single ride 1 and single ride 2 in Fig.5(a) start from and end at the same grids as the shared ride 1 and shared ride 2 do, respectively. In this case, the two single rides in Fig.5(a) are regarded as potential substitutes for shared rides in Fig.5(b). To be more representative, all the potential substituted single rides with origins and destinations in the same grids are selected to calculate the median trip distance and $CO_2$ emissions of the substituted single rides, as shown in Eq (11)-(12). Then these median values are used as baselines (minuends) to estimate the travel distance saved and $CO_2$ emission reduction by each ridesplitting trip $s$, as shown in Eq (13)-(14). For better comparison, the percentage of $CO_2$ emission reduction from substituted single rides and $CO_2$ emission reduction per kilometer of the ridesplitting trip are calculated as Eq (15)-(16).

$$TD^n = \text{median}\{TD^k\}, \quad k \in \{k \mid OG^k = OG^n \ \& \ DG^k = DG^n\} \tag{11}$$

$$E^n = \text{median}\{E^k\}, \quad k \in \{k \mid OG^k = OG^n \ \& \ DG^k = DG^n\} \tag{12}$$

$$SD^s = \sum_n^{NSR} TD^n - TD^s \tag{13}$$

$$ER^s = \sum_n^{NSR} E^n - E^s \tag{14}$$

$$ERP^s = ER^s / \sum_n^{NSR} E^n \tag{15}$$

$$ERR^s = ER^s / TD^s \tag{16}$$

where $TD^n$ is the median (baseline) trip distance of substituted single rides for shared ride $n$;

*OG* denotes the origin grid; *DG* denotes the destination grid; $E^n$ is the median (baseline) $CO_2$ emission of substituted single rides for shared ride *n*; $SD^s$ and $ER^s$ are the travel distance saved and $CO_2$ emission reduction by ridesplitting trip *s* compared with substituted single rides, respectively; $ERP^s$ and $ERR^s$ are $CO_2$ emission reduction percentage (%) and rate (g/km) of ridesplitting trip *s*.

**4.3 Calculation of explanatory variables of emission reduction**

To further analyze the influencing factors of the $CO_2$ emission reduction of ridesplitting trips, some possible explanatory variables are also calculated based on the data above. The first factor is the overlap distance of shared rides, i.e., the distance of trip trajectory where the riders shared with others in the same ridesplitting trip. It can be calculated by summing the travel distances of the trajectory segments during the overlap time of all shared rides, as shown in Eq (17). Then the overlap rate, which is the proportion of overlap distance in the ridesplitting trip distance, can be calculated as Eq (18). It reflects the trajectory similarity of multiple shared rides in the ridesplitting trip.

$$OD^s = \sum_{i \in \Phi^s} d_i^s, \quad \Phi^s \equiv \{i \mid OT_{start}^s \leq ts_i^s \leq OT_{end}^s\} \tag{17}$$

$$OR^s = OD^s / TD^s \tag{18}$$

where $OD^s$ is the overlap distance of shared rides in each ridesplitting trip *s*; $\Phi^s$ is the finite set of overlapping trajectory segment *i* of ridesplitting trip *s*; $ts_i^s$ is the timestamp of segment *i* of ridesplitting trip *s*; $OT_{start}^s$ and $OT_{end}^s$ is the start and end times of overlapping trajectory of ridesplitting trip *s*, respectively; $OR^s$ is the overlap rate of ridesplitting trip *s*.

The second factor is the detour distance of the ridesplitting trip, i.e., the additional travel distance of riders to pick up or drop off others sharing in the same trip. It can be obtained by comparing the actual trip distance of the shared ride *n* with the median trip distance of corresponding substituted single rides and summing the differences for each shared ride *n*, as shown in Eq (19). The actual trip distance of the shared ride *n* is calculated and calibrated as Eq (20). Then, the detour rate, which is the proportion of detour distance in the ridesplitting trip distance, can be calculated as Eq (21) to reflect the negative effects of ridesplitting.

$$DD^s = \sum_n^{NSR} \left(TD^n - \overline{TD^n}\right) \tag{19}$$

$$TD^n = C \times \sum_{i \in \Omega^n} d_i^n, \quad \Omega^n \equiv \{i \mid ts_{start}^n \leq ts_i^n \leq ts_{end}^n\} \tag{20}$$

$$DR^s = DD^s / TD^s \qquad (21)$$

where $DD^s$ is the detour distance of ridesplitting trip *s*; $TD^n$ is the actual trip distance of shared ride *n* in ridesplitting trip *s*; $\Omega^n$ is the finite set of trajectory segment *i* of the shared ride *n*; $ts_i^n$ is the timestamp of segment *i* of the shared ride *n*; $ts_{start}^n$ and $ts_{end}^n$ is the start and end timestamps of the shared ride *n*, respectively; $DR^s$ is the detour rate of ridesplitting trip *s*.

Other trip-related variables such as the number of shared rides, start hour, average speed, the actual trip distance of the ridesplitting trip, and minimum, maximum, and total trip distance of substituted single rides for shared rides in a ridesplitting trip can be easily calculated based on trip characteristics above. The calculation equations of these variables are omitted here, and a detailed description of all the variables is presented in Table 3.

## 4.4 Machine learning models for emission reduction regression

### 4.4.1 Gradient boosting machines

To model the relationship between the dependent variable (i.e., $CO_2$ emission reduction rate of ridesplitting) and the above explanatory variables, the Gradient Boosting Machines (GBM) are employed for regression. The GBM is a powerful machine learning algorithm that gives a prediction model in the form of an ensemble of decision trees, with each tree learning and improving on the previous ones [35]. It is proven to outperform the Bagging algorithms, such as random forest, on the prediction precision [36]. There are four popular implementations of the gradient boosting framework, i.e., Gradient-Boosted Decision Trees (GBDT), eXtreme Gradient Boosting (XGBoost), LightGBM, and CatBoost. This study applies the four algorithms to predict the $CO_2$ emission reduction rate of ridesplitting and compares their performances. The brief introductions of these four models are presented as follows:

**(1) GBDT model**

The GBDT was proposed by Friedman [37] in 2001. It is the basic form of GBM based on iterative decision trees. It starts with a weak learner (e.g., a decision tree with only a few splits) and sequentially boosts its performance by fitting into the residuals of the previous trees. In each iteration, the negative gradient of the loss function is found so that the loss function can be decreased as quickly as possible to achieve the goal of minimizing the loss function. The final model aggregates the result of each weak learner, and thus, a strong learner is obtained.

**(2) XGBoost model**

The XGBoost model was proposed by Chen and Guestrin [38]. It is an efficient

implementation of the GBDT. There are two improvements on XGBoost compared with GBDT. First, it introduces a regularization term into the objective function to prevent overfitting. Second, it computes second-order gradients, i.e., second partial derivatives of the loss function, which provide more information about the direction of gradients and how to get to the minimum of the loss function. Therefore, the XGBoost model can significantly improve the prediction accuracy and training speed of the GBDT model.

**(3) LightGBM model**

The LightGBM model was developed by Microsoft as a fast and scalable alternative to XGBoost and GBDT [39]. To reduce memory usage and computing time, it uses histogram-based algorithms, which bucket continuous feature values into discrete bins. Instead of using trees level (depth)-wise strategy as most decision tree learning algorithms do, it grows trees leaf-wise (best-first) and splits the leaf with max information gain. In general, leaf-wise algorithms tend to achieve lower loss than level-wise algorithms. In addition, it utilizes two novel techniques called Gradient-Based One-Side Sampling (GOSS) and Exclusive Feature Bundling (EFB) to reduce the number of instances and features [39]. Therefore, these features enable the LightGBM model to run faster while maintaining a high level of accuracy.

**(4) CatBoost model**

The Catboost model was developed by Yandex researchers and engineers in 2017 [40]. Two main innovations lead to the CatBoost model outperforming other implementations of GBM. First, it uses the concept of ordered boosting, a permutation-driven approach to train the model on a subset of data while calculating residuals on another subset, thus preventing target leakage and overfitting. Second, it provides native (automatic) handling for categorical features to avoid the pre-processing of data. Both techniques make CatBoost able to solve the problem of prediction shift caused by a special kind of target leakage.

*4.4.2 Model interpretation methods*

Although some advanced machine learning algorithms have great potential for improving the prediction accuracy of the outputs, they are sometimes defined as black boxes compared with interpretable models such as linear regression. To better understand the effect of each feature on the output, the following two interpretation methods are used to explain the results of the GBMs mentioned above.

**(1) Partial Dependence Plot**

The partial dependence plot (PDP) shows the marginal effect that one or two features have on the target of a machine learning model. The PDP with one feature can indicate whether the relationship between the target and the feature is linear, monotonic, or more complex. Moreover, the PDP with two features shows the interactions between the target response and joint effects of the two features. The function of PDP can be defined as the expectation of model outputs depending only on the specific features of interest, as shown in Eq (22). It is usually estimated by calculating averages in the training data as Eq (23).

$$\hat{f}_S(x_S) = E_{X_C}\left[f(x_S, X_C)\right] = \int f(x_S, x_C) p(x_C) dx_C \tag{22}$$

$$\hat{f}_S(x_S) \approx \frac{1}{M} \sum_{m=1}^{M} f(x_S, x_C^{(m)}) \tag{23}$$

where $f$ defines the function of the machine learning model, $x_S$ denotes the one or two features of interest, $X_C$ is the set of other features $x_C$ used in the model, $\hat{f}_S(x_S)$ is the partial dependence function for regression at a point $x_S$, $f(x_S, x_C^{(m)})$ is the model prediction for a specific $m$-th sample whose feature values are determined by $x_S$ and $x_C$, and $M$ is the number of samples.

To sum up, PDP is a global method that considers all samples and describes the global relationship between the features of interest and the predicted outcome. Given the PDP, we may further analyze the relationships between the $CO_2$ emission reduction rate of ridesplitting and the explanatory variables. However, it should be noted that the PDP assumes that the features of interest are not correlated with the complement features. Otherwise, some absurd data points will be computed to create the PDP. Therefore, the feature selection should be careful to avoid the multicollinearity of the explanatory variables.

**(2)  SHapley Additive exPlanations**

The SHapley Additive exPlanations (SHAP), proposed by Lundberg and Lee [41], is a unified method to interpret model predictions and measure the feature importance. It is based on the Shapley value, which is a widely used approach from cooperative game theory [42]. The goal of SHAP is to explain the prediction of a specific sample by calculating the contribution of each feature to the prediction. The explanation model is a linear function of binary variables, as shown in Eq (24).

$$g(x') = \phi_0 + \sum_{j=1}^{J} \phi_j x'_j \tag{24}$$

where $g(x')$ is the explanation model for a specific input $x$, $x' \in \{0,1\}^J$ is the vector of simplified features that map to the original features, $J$ is the number of simplified features, $\phi_j$ is the effect of the *j*-th feature on the model output, and $\phi_0$ is the model output with all simplified features excluded. If we set $\phi_0$ the expected value of all predictions and set $x'_j$ to 1, the explanation model is approximated to the original model *f*, as shown in Eq (25). It reflects the local accuracy of the SHAP method.

$$f(x) = g(x') = E_X\left[f(X)\right] + \sum_{j=1}^{J} \phi_j \qquad (25)$$

For a particular single prediction, $E_X\left[f(X)\right]$ is regarded as a baseline (expected) model output, $\phi_j$ is defined as the SHAP value, which presents the contribution of the *j*-th feature to the prediction. The SHAP value is computed by the difference in model predictions of including and excluding that feature. Since the effect of withholding a feature depends on other features in the model, the preceding differences are weighted and summed over all possible feature value combinations, as shown in Eq (26).

$$\phi_j = \sum_{S \subseteq F \setminus \{j\}} \frac{|S|!(|F|-|S|-1)!}{|F|!}\left[f_{S \cup \{j\}}(x_{S \cup \{j\}}) - f_S(x_S)\right] \qquad (26)$$

where *F* is the set of all features used in the model and $|F|$ is the size of set *F*, *S* is any subset of features without the *j*-th feature and $|S|$ is the size of set *S*, $f_S(x_S)$ is the model prediction for features in set *S* that are marginalized over features not included in set *S*.

As SHAP values are very complicated to compute (they are NP-hard in general), some approximation methods are usually used to estimate the SHAP values, such as Monte-Carlo sampling, Kernel SHAP, and TreeSHAP [43]. Given the SHAP values of each feature across the data, we can obtain the SHAP feature importance, which presents the global contribution of the feature, as shown in Eq (27).

$$I_j = \frac{1}{M}\sum_{m=1}^{M}\left|\phi_j^{(m)}\right| \qquad (27)$$

where $I_j$ is the global importance value of the *j*-th feature, $\phi_j^{(m)}$ is the SHAP value of the *j*-th feature for the *m*-th sample.

## 5 Results and discussions

### 5.1 Characteristics of CO$_2$ emission reduction of ridesplitting

Based on the data and the method above, all the variables related to the CO$_2$ emission reduction of ridesplitting are calculated, as shown in Table 3. The abnormal ridesplitting trips with *Overlap distance* less than 500 m and *Overlap time* less than 1 min are filtered because contiguous single rides may be mistaken as shared rides. In addition, the ridesplitting trips with outliers of variables are also filtered based on the 1.5 times Interquartile Range (IQR) method. It should be noted that the ridesplitting trips with 4 shared rides are filtered because of the small sample size and feature instability. In this case, only 16236 valid samples of ridesplitting trips remain for further analysis.

Table 3  Definition and descriptive analysis of the variables.

| Variables | Description | Mean | Std | Min | Max |
|---|---|---|---|---|---|
| *Saved distance* (km) | Travel distance saved by the ridesplitting trip compared with substituted single rides | 2.11 | 1.89 | -4.26 | 17.59 |
| *Emission reduction* (g) | $CO_2$ emission reduction of the ridesplitting trip compared with substituted single rides | 307.23 | 324.79 | -719.92 | 2754.99 |
| *Emission reduction percentage* (%) | Percentage of $CO_2$ emission reduction from substituted single rides | 16.49 | 15.58 | -27.26 | 59.83 |
| *Emission reduction rate* (g/km) | $CO_2$ emission reduction per kilometer of the ridesplitting trip | 43.15 | 43.01 | -45.62 | 271.82 |
| *Overlap distance* (km) | Overlap distance of shared rides in the ridesplitting trip | 3.65 | 1.81 | 0.50 | 15.92 |
| *Overlap rate* | The proportion of overlap distance in the ridesplitting trip distance | 0.48 | 0.18 | 0.05 | 1.00 |
| *Detour distance* (km) | Detour distance of the ridespltting trip | 1.74 | 1.88 | -2.91 | 15.91 |
| *Detour rate* | The proportion of detour distance in the ridesplitting trip distance | 0.20 | 0.18 | -0.31 | 0.72 |
| *Number of shared rides* | Number of shared rides in the ridesplitting trip | 2.04 | 0.19 | 2.00 | 3.00 |
| *Peak hours* (0 or 1) | Whether the ridesplitting trip starts at peak hours (7-9 am and 5-8 pm on weekdays) | 0.24 | 0.43 | 0.00 | 1.00 |
| *Average speed* (km/h) | Average speed of the ridespltting trip | 17.77 | 4.04 | 6.20 | 45.32 |
| *Actual trip distance* (km) | Actual trip distance of the ridesplitting trip | 8.03 | 2.77 | 1.59 | 22.77 |
| *Min ride distance* (km) | Minimum trip distance of substituted single rides | 3.99 | 1.59 | 0.81 | 11.39 |
| *Max ride distance* (km) | Maximum trip distance of substituted single rides | 5.96 | 2.01 | 1.10 | 17.84 |
| *Total ride distance* (km) | Sum of trip distances of substituted single rides | 10.14 | 3.47 | 2.18 | 32.45 |
| *Ride distance gap* (km) | Gap between min ride distance and max ride distance | 1.97 | 1.64 | 0.00 | 14.09 |
| *Ride distance ratio* | Ratio between min ride distance and max ride distance | 0.69 | 0.20 | 0.11 | 1.00 |

The results show that the average saved distance of ridesplitting trips is 2.11 km; the average emission reduction of ridesplitting trips is 307.23g; the average emission reduction percentage is 16.49%; the average emission reduction rate is 43.15g/km. Fig. 6 presents a linear correlation between travel distances saved and $CO_2$ emission reduction by ridesplitting trips with an R-squared of 0.908. It indicates that the emission reduction of ridesplitting is directly related to the travel distance saved by ridesplitting. In addition, the histograms of distance saved per trip and $CO_2$ emission reduction per trip are also plotted on the top x-axis and right y-axis in Fig. 6, respectively. We are surprised to find that about 10% of ridesplitting trips failed to save distance, and 15% of ridesplitting trips failed to reduce $CO_2$ emissions. This is probably due to the high rate of detours for picking up and dropping off other shared riders so that the detour rate could be even up to 0.72.

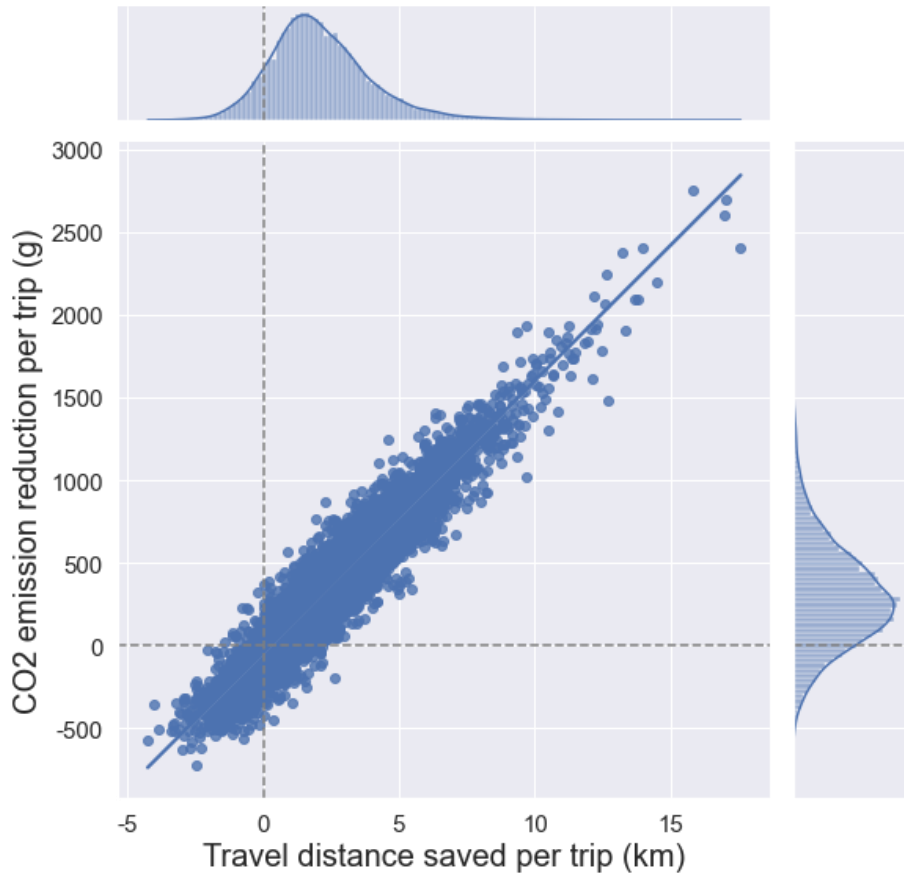

**Fig. 6 The relationship between distances saved and $CO_2$ emission reductions by ridesplitting trips.**

Furthermore, the $CO_2$ emission reductions of the ridesplitting trips are grouped by 24 hours of every day. The temporal distributions of $CO_2$ emission reduction of ridesplitting trips with both 2 and 3 shared rides are presented in Fig. 7. It shows that the hourly $CO_2$ emission reductions of ridesplitting trips are more concentrated during 13:00-15:00 and 22:00-23:00. While the $CO_2$ emission reduction rate of ridesplitting is especially higher in the early morning. It indicates that ridesplitting may have more potential to reduce $CO_2$ emissions during non-peak hours. Comparing the ridesplitting trips with 2 and 3 shared rides, the former contributes most of the total $CO_2$ emission reduction of ridesplitting trips. In contrast, the $CO_2$ emission reduction rate of ridesplitting trips with 3 shared rides is significantly higher.

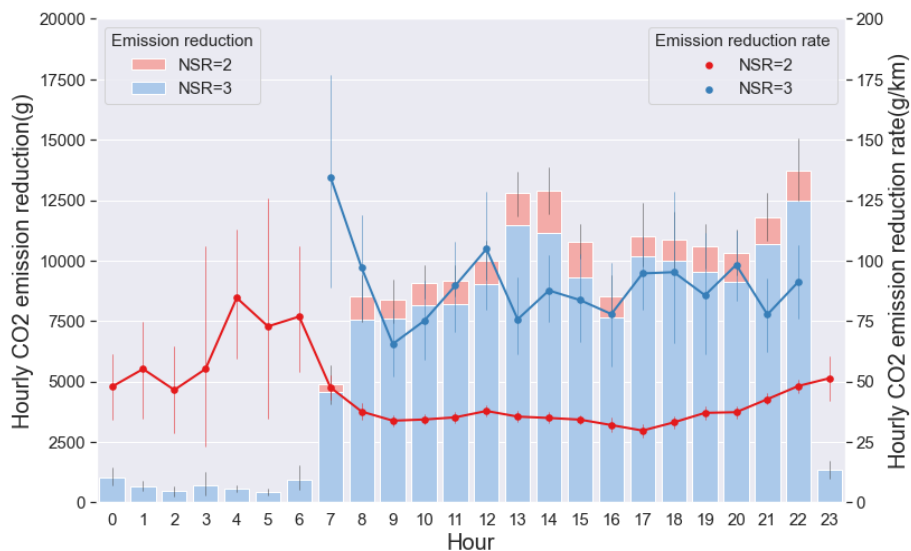

**Fig. 7 Temporal distributions of CO$_2$ emission reduction of ridesplitting trips.**

Since the travel distances saved and CO$_2$ emission reductions by ridesplitting trips highly depend on trip distance, we only focus on the CO$_2$ emission reduction per kilometer for further analysis. To present multiple distributions of CO$_2$ emission reduction rate of ridesplitting across the number of shared rides and start hours, a violin plot that plays a similar role as a box plot is presented in Fig. 8. The results show that increasing the number of shared rides from 2 to 3 can significantly increase the average CO$_2$ emission reduction rate of ridesplitting from 41.41 to 89.94 g/km. On the other hand, the distributions of the CO$_2$ emission reduction rate of ridesplitting in peak hours and non-peak hours are different, especially for those with 3 shared rides. It also indicates that traveling in peak hours may lower the CO$_2$ emission reduction rate of ridesplitting trips.

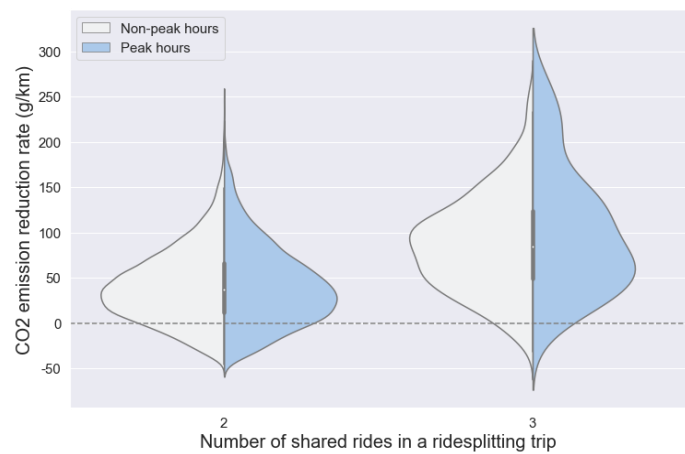

**Fig. 8 Distributions of CO$_2$ emission reduction rate of ridesplitting across the number of shared rides and start hours.**

## 5.2 Regressions of $CO_2$ emission reduction rate of ridesplitting

Before the regression, pairwise correlations of all variables related to $CO_2$ emission reductions of ridesplitting are calculated. A heatmap is used to visualize the correlation matrix, as shown in Fig 9. It can be found that the *Saved distance*, *Emission reduction*, *Emission reduction percentage,* and *Emission reduction rate* are significantly correlated to each other, which represent the benefits of the ridesplitting. To eliminate the scaling effects, only the *Emission reduction rate* is selected as the dependent variable of regression. Meanwhile, considering the correlation and multicollinearity of variables, the *Overlap rate*, *Detour rate*, *Number of shared rides*, *Peak hours*, *Average speed*, *Actual trip distance*, *Ride distance gap*, and *Ride distance ratio* are selected as independent variables for regression.

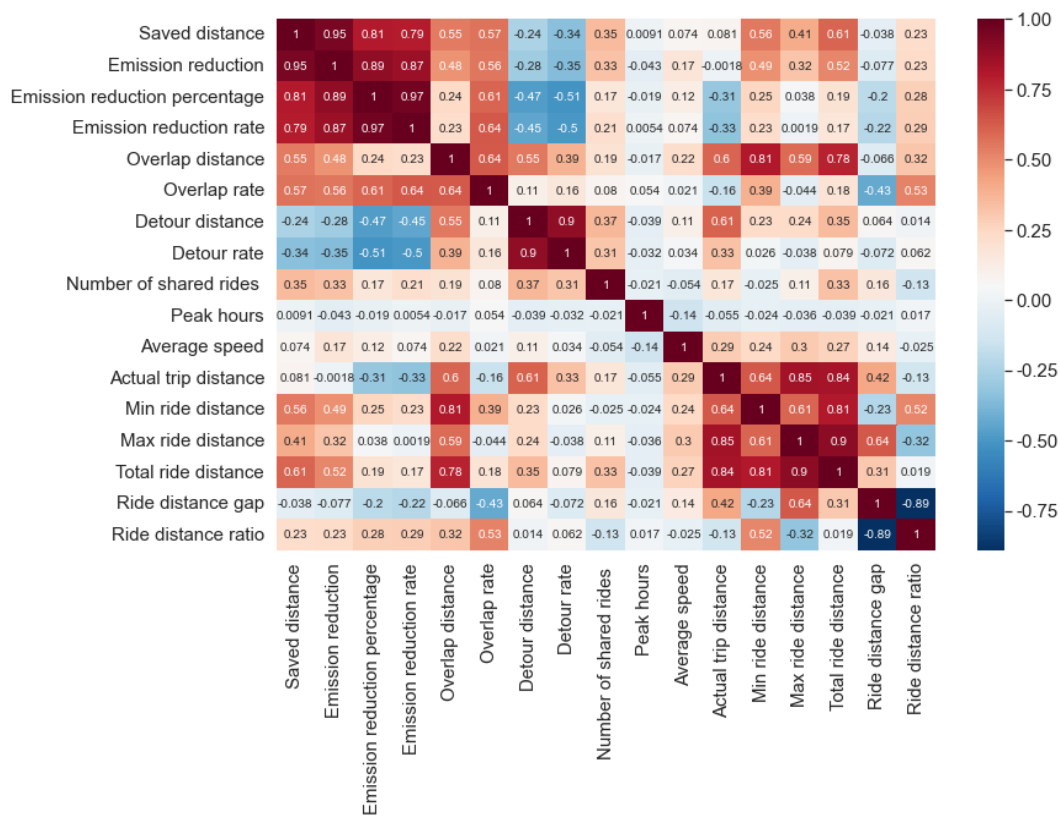

**Fig. 9 Correlation heatmap of all variables related to ridesplitting.**

The data of the selected variables are split into the training set and test set with a ratio of 8:2. Then, the GBDT, XGboost, LighGBM, and CatBoost models are trained on the training set. To obtain high accuracy and avoid overfitting of the prediction, three critical hyper-parameters are tuned for the GBMs, including iterations (i.e., the maximum number of trees that be boosted), learning rate (i.e., shrinkage), and depth (i.e., maximum depth of each tree). For the iterations, we examine values from 2000 to 5000 at an interval of 500. For the learning rate, we examine 0.005, 0.01, and 0.05. For the depth, we examine values from 3 to 8 at an interval of 1. The rest of the other hyper-parameters are set to default values. In total, 126 possible combinations of hyper-parameters are tested using five-fold cross-validation and grid-search methods for best models. Finally, the root mean squared error (RMSE), mean absolute error (MAE), and $R^2$ are used to evaluate the performances of the best models on the test set for the four implementations of GBMs. The best hyper-parameters and evaluation metrics are shown in Table 4. The traditional linear regression model is also fitted using the Ordinary Least Squares (OLS) method for comparison. The results indicate that the CatBoost model has the best performance among the five models. Therefore, only the trained CatBoost model is used in the following analysis.

**Table 4 Model performance evaluation and comparison**

| Models | Best hyper-parameters | | | Performance evaluation on the test set | | |
|---|---|---|---|---|---|---|
| | Iterations | Learning rate | Depth | RMSE | MAE | $R^2$ |
| OLS | - | - | - | 11.8015 | 8.7572 | 0.928 |
| GBDT | 4500 | 0.005 | 3 | 10.7224 | 8.2358 | 0.9386 |
| XGBoost | 2500 | 0.01 | 3 | 10.7227 | 8.2333 | 0.9386 |
| LightGBM | 3500 | 0.01 | 3 | 10.7117 | 8.2195 | 0.9387 |
| **CatBoost** | **4500** | **0.005** | **4** | **10.4561** | **8.0109** | **0.9416** |

### 5.3 Determinants of CO2 emission reduction rate of ridesplitting

To interpret the trained CatBoost model, the SHAP value of each feature for each sample can be computed based on Eq (26), as shown in Fig.10(a). It presents the distribution of the impacts each feature has on model output (the $CO_2$ emission reduction rate of ridesplitting). The color represents the feature value from low (blue) to high (red). The distribution of feature values across the SHAP values reveals the impact direction of features on the model output. For example, higher *Overlap rate*, *Number of shared rides*, *Average speed*, and *Ride distance ratio* lead to a higher $CO_2$ emission reduction rate of ridesplitting. On the contrary, an increase in the *Detour rate*, *Actual trip distance*, *Peak hours*, and *Ride distance gap* has negative effects on the $CO_2$ emission reduction rate of ridesplitting. By computing the mean absolute SHAP values of all samples for each feature, we can obtain a global measure of feature importance based on Eq (27). Then the features are sorted in descending order by the feature importance, as shown in Fig.10(b). It indicates that the *Overlap rate* and *Detour rate* are the most important determinants of the $CO_2$ emission reduction rate of ridesplitting, accounting for 77% of the average impact on model output, followed by the *Number of shared rides*, *Average speed*, *Actual trip distance*, *Peak hours*, *Ride distance gap*, and *Ride distance ratio*.

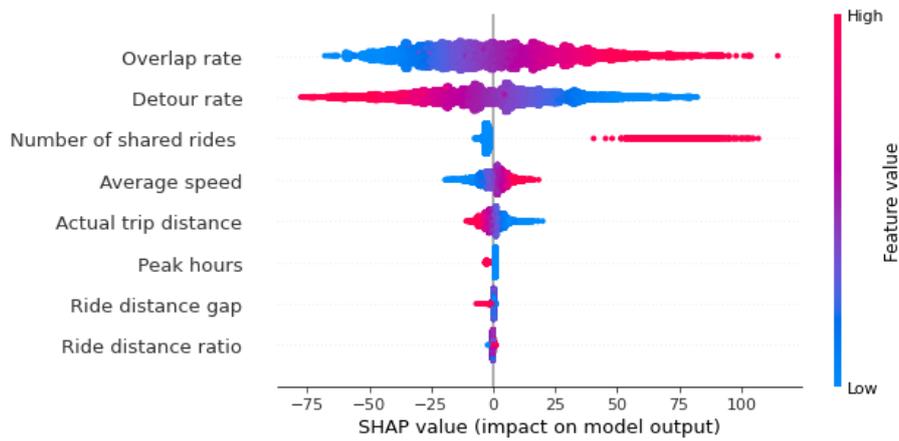

**(a)** Distribution of SHAP values for each feature

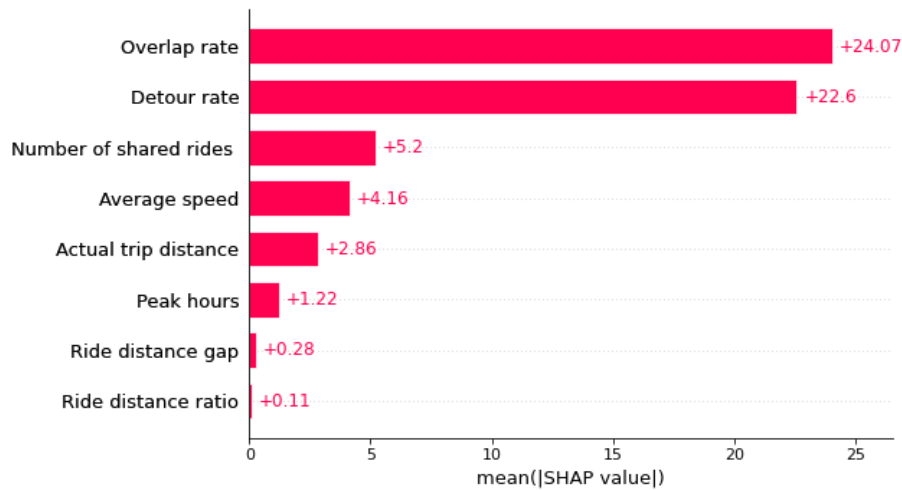

**(b)** Mean absolute SHAP values of each feature

**Fig. 10 Feature importance based on SHAP values.**

The PDP of each feature is used to visualize the quantitative relationships between the $CO_2$ emission reduction rate of ridesplitting and its determinants, as shown in Fig. 11. It represents the average marginal effect of each feature on the model outputs. Therefore, the causality between the feature and the model outputs can be interpreted based on the PDPs. It should be noted that the relationship is only causal for the model but not necessarily for the real world. The histogram of each feature is also plotted on the x-axis to present the distribution of samples across different feature values. If the feature values are in the ranges with a lower frequency of samples, the interpretation may be less reliable.

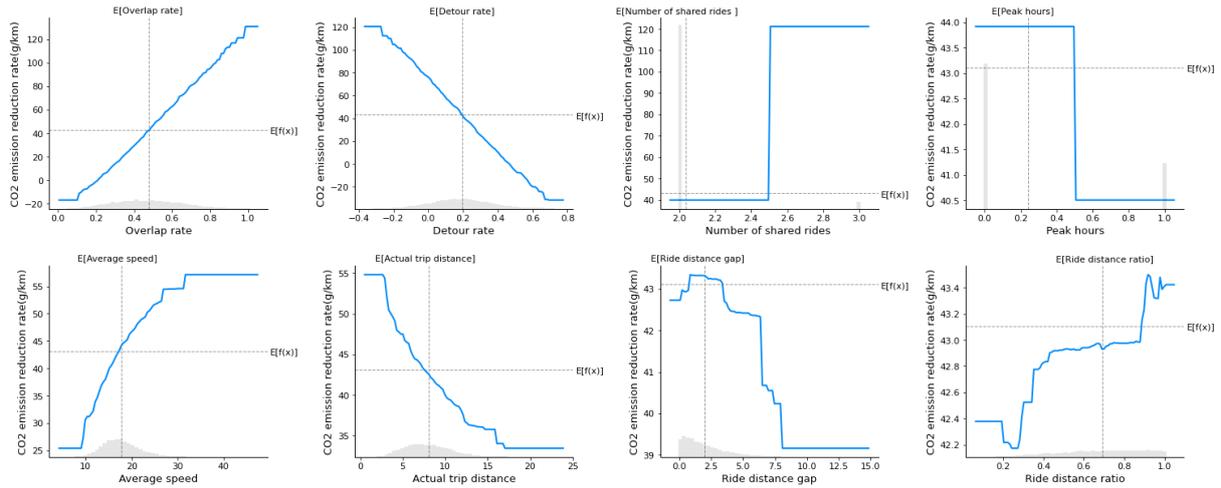

Fig. 11 Partial dependence plot of each feature

The average effect trends of these features in Fig. 11 are consistent with the indications of the relationships in Fig. 10(a). The *Overlap rate* and *Detour rate* seem to be linearly correlated to the $CO_2$ emission reduction rate of ridesplitting and cause the largest changes in the model output. If the *Overlap rate* rises 0.1, the model output will be raised by about 15 g/km on average. On the contrary, if the *Detour rate* rises 0.1, the model output will decrease by about 15 g/km on average. What is more, increasing the number of shared rides from 2 to 3 results in an increase in model output from 40 to 120 g/km on average. When a ridesplitting trip starts in peak hours, the $CO_2$ emission reduction rate of ridesplitting may reduce by 4 g/km on average. The relationships between the $CO_2$ emission reduction rate of ridesplitting and other features are nonlinear. With the increase in *Average speed* and *Actual trip distance*, the gradients of PDP curves decrease, which indicates that the effects of these two features are weakening. In addition, there are some salient thresholds for the effects of the *Ride distance gap* and *Ride distance ratio*. For example, the $CO_2$ emission reduction rate of ridesplitting decreases significantly when the *Ride distance gap* increase from 3.5 km to 7.5 km; while it increases significantly when the *Ride distance ratio* increase from 0.3 and 0.45. These thresholds of features are useful for operators to expand the environmental benefits of ridesplitting.

The PDPs with two features can represent the interactions among the two features. Since the *Overlap rate* is the most important feature, the interaction effects of the *Overlap rate* and any other feature on the $CO_2$ emission reduction rate of ridesplitting are shown in Fig.12. The x-axis and y-axis are the values of two features of interest, the color-filled contours represent the dependence of model output on the joint values of the two features. The values of model output are classified into several contour levels, from low (red) to high (blue). If the pair of feature values are in the region of blue contours, then the ridesplitting trip can significantly reduce $CO_2$ emissions. Otherwise, the emission reduction rate of the ridesplitting trip is low or even negative. To be specific, when the *Detour rate* is smaller than 0.07 or the *Number of shared rides* is 3, ridesplitting can reduce $CO_2$ emission (model output>0) regardless of the *Overlap rate*. With the decrease in *Average speed* and increase in *Actual trip distance*, the *Overlap rate* needs to be higher to ensure ridesplitting to reduce emissions (model output > 0). Therefore, the PDPs with two features can provide decision boundaries for operators to match the shared rides that have greater potential to reduce $CO_2$ emissions.

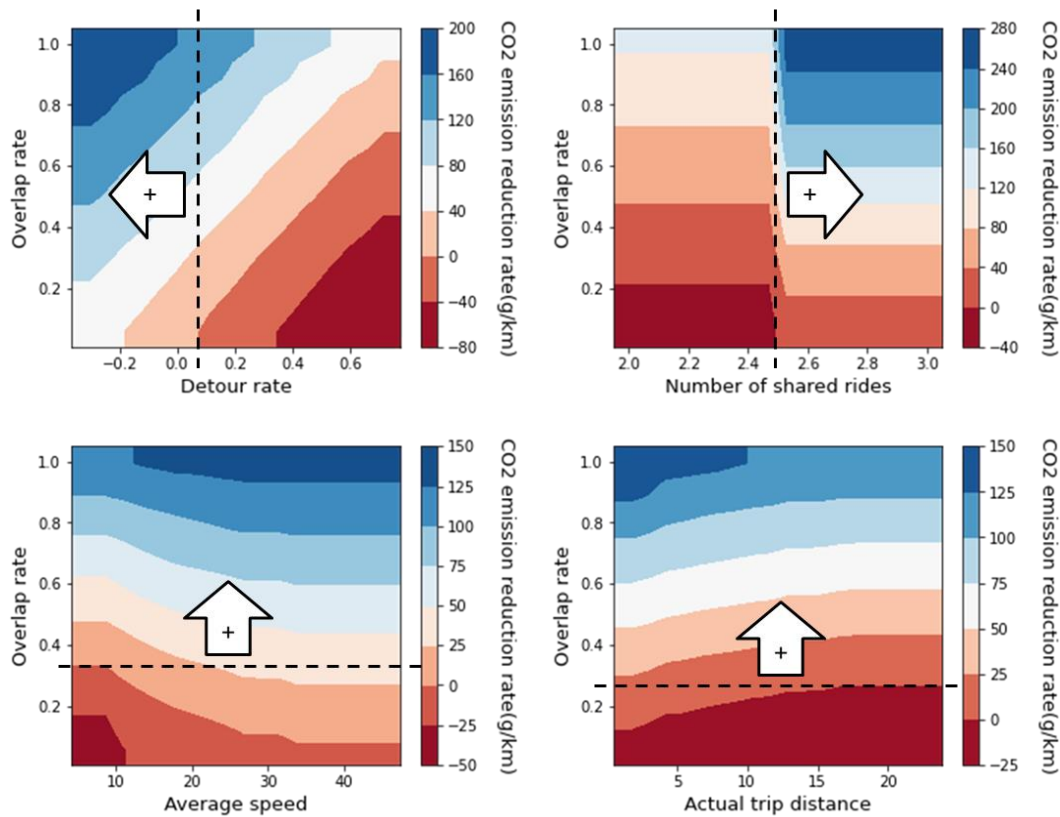

**Fig. 12 Interaction of overlap rate and any other feature on the output.**

The SHAP dependence plot is an alternative to PDP. While PDP shows average effects, SHAP dependence also shows the variance of model output on the y-axis at a single feature value, which is useful to highlight feature interactions [44]. Therefore, we can also visualize the feature dependence with the interaction effect based on the SHAP values, as shown in Fig. 13. Since SHAP values represent a feature's responsibility for a change in the model output, the plots below represent the change in the predicted $CO_2$ emission reduction rate of ridesplitting as each feature changes interacting with the number of shared rides. To help reveal these interactions, the SHAP dependence plot for ridesplitting trips with the 2 and 3 shared rides are colored in blue and red, respectively. It shows that the number of shared rides may influence the individual effect of each other features on the model output. When increasing the number of shared rides in ridesplitting trips, the positive effects of the *Overlap rate* and *Ride distance ratio* are enhanced; while the negative effects of the *Detour rate*, *Peak hour*, and *Actual trip distance* are weakened. Therefore, the influence patterns of features on the $CO_2$ emission reduction rate of ridesplitting are significantly different for ridesplitting trips with different numbers of shared rides.

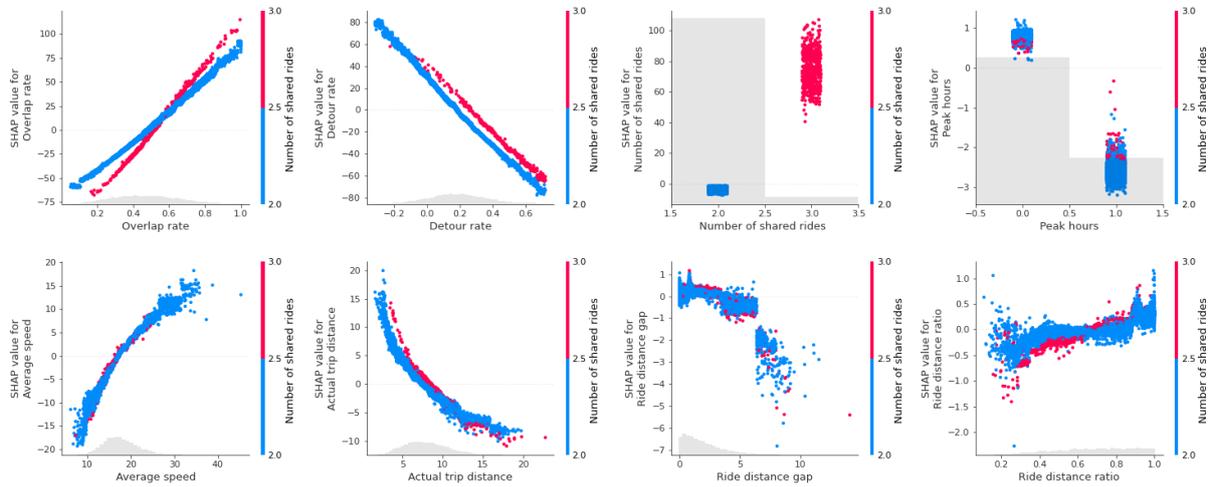

**Fig. 13 SHAP feature dependence plots with interactions visualization.**

## 6 Conclusions and implications

To reveal what determines the environmental benefits of ridesplitting trips in the real world, this study explores the $CO_2$ emission reduction of ridesplitting and its determinants based on the order and trajectory data of ridesourcing services of Didi Chuxing in Chengdu. First, the trip characteristics of both single rides and shared rides are extracted from the datasets. Second, the $CO_2$ emission of each trip is calculated based on the COPERT model. Then, the $CO_2$ emission reduction of ridesplitting is estimated by comparing the $CO_2$ emissions of shared rides and their substituted single rides with the same grids of origins and destinations. Finally, the relationship between the $CO_2$ emission reduction rate of ridesplitting and its explanatory variables are analyzed using four implementations of gradient boosting machines, including GBDT, XGboost, LighGBM, and CatBoost models. The main findings are concluded as follows:

(1) Compared with regular ridesourcing, a ridesplitting trip can averagely save travel distance by 2.11 km and reduce $CO_2$ emission by 307.23 g accounting for 16.49% of total emissions of its substituted single rides. The average $CO_2$ emission reduction rate of ridesplitting is 43.15g/km.

(2) Not all ridesplitting trips reduce emissions from ridesourcing in the real world. Due to

the possible detours for picking up and dropping off additional shared riders, about 10% and 15% of ridesplitting trips may even increase travel distances and $CO_2$ emissions, respectively.

(3) The $CO_2$ emission reduction rate of ridesplitting varies from trip to trip depending on the attributes of shared rides and their substituted single rides. It is positively associated with the *Overlap rate*, *Number of shared rides, Average speed*, and *Ride distance ratio* can increase, whereas it is negatively associated with *Detour rate*, *Peak hours*, *Actual trip distance*, and *Ride distance gap.*

(4) The *Overlap rate* and *Detour rate* are the most important determinants, changing the $CO_2$ emission reduction rate of ridesplitting on average by +15 g/km and -15 g/km with an increase of 0.1 in *Overlap rate* and *Detour rate*, respectively.

(5) Nonlinear effects and interaction effects are observed in the relationship between the $CO_2$ emission reduction rate of ridesplitting and its determinants, which provides a useful reference for the decision-making of TNCs to match shared rides.

This study can provide a scientific method for TNCs to better evaluate and optimize the environmental benefits of ridesplitting and help the government to introduce policies to promote ridesplitting. More specifically, some implications for the government and TNCs are summarized as follows:

(1) Based on the CatBoost model trained in this study, TNCs can easily predict the $CO_2$ emission reduction of ridesplitting without a complex calculation of the $CO_2$ emission of ridesplitting. Therefore, the shared rides that fail to reduce travel distance and emissions should be avoided to be matched into a ridesplitting trip.

(2) The current average $CO_2$ emission reduction percentage of ridesplitting is still low (16.49%). TNCs should further increase the *Overlap rate* and decrease the *Detour rate* of shared rides by optimizing their algorithms of ride assignment and vehicle routing.

(3) The ridesplitting trips shared by more than 3 riders have much greater potential to reduce $CO_2$ emissions. The government and TNCs could enact some incentive policies to encourage more people to adopt ridesplitting and increase the passenger capacities of ridesourcing vehicles.

(4) When the *Detour rate* is larger, the *Number of shared rides* is smaller, the *Average speed* is lower, and the *Travel distance* is longer, TNCs should further increase the *Overlap rate* to ensure $CO_2$ emission reduction of ridesplitting.

(5) Given the observed data and the method proposed in this study, the carbon footprint and the $CO_2$ emission reduction of each ridesplitting trip could be accurately calculated. It enables the government or TNCs to issue carbon credits to travelers for a low-carbon transition of urban transport [45].

However, this study also has some limitations, which should be addressed in the future. First, this study only calculates the direct emission reduction of ridesplitting using the substituted single rides as the baseline. The indirect emission reduction resulting from the changes in behavioral habits and car ownership in the long term should be further assessed. Second, when calculating the emissions of ridesourcing trips, this study assumes all the vehicles are petrol cars complying with the same emission standard. With the rapid adoption of electric cars in ridesourcing services, the impact of electric cars should also be considered in the environmental benefit assessment. Third, the findings of this study may only reveal the evidence in the specific area of Chengdu. Based on the research framework proposed in this study, more empirical studies of other cities could be conducted to further validate the results of this study.

## Acknowledgements

This study was sponsored by the National Natural Science Foundation of China [Grant number: 52002244]; the Chenguang Program supported by Shanghai Education Development Foundation and Shanghai Municipal Education Commission [Grant number: 20CG55]; and the Shanghai Pujiang Program [Grant number: 2020PJC083].